\documentclass[11pt]{article}

\usepackage{acl}

\usepackage{times}
\usepackage{latexsym}

\usepackage[T1]{fontenc}

\usepackage[utf8]{inputenc}

\usepackage{microtype}

\usepackage{inconsolata}

\usepackage{graphicx}
\usepackage{amsmath}
\usepackage{amssymb}
\usepackage{algorithm}
\usepackage{algorithmic}
\usepackage{booktabs}
\usepackage{multirow}

\usepackage[table]{xcolor}
\title{TopoCompress:\\Long Context Compression via Graph-Wired Semantic Trajectories}

\author{
Daniel Agyei Asante\textsuperscript{1},
Yang Li\textsuperscript{1}\thanks{
 Corresponding author. 
 Email: \href{mailto:jerryyangli@gmail.com}{jerryyangli@gmail.com}.
 } \\
\textsuperscript{1}Iowa State University, United States \\
{\tt \{dasante, yangli1\}@iastate.edu}
}

\begin{document}
\maketitle
 \begin{abstract}
Long-context compression is essential for reducing the cost and latency of large language model inference. However, existing methods can fragment important evidence, require additional training or alignment, and often depend on the target model for effective compression. We introduce \texttt{TopoCompress}, a training-free and model-agnostic framework that compresses long contexts by selecting coherent semantic spans. \texttt{TopoCompress} first scores each span using dense and lexical query relevance together with semantic acceleration. It then constructs a hybrid graph that connects spans based on semantic similarity and sequential adjacency, and propagates the query-guided relevance scores over the graph.  Across five long-context tasks—HotpotQA, 2WikiMQA, MuSiQue, Qasper, and MultiFieldQA-en—\texttt{TopoCompress} consistently outperforms strong compression baselines. Notably, \texttt{TopoCompress} achieves performance comparable to the strongest baseline while using a $4\times$ smaller compression budget, and provides a $1.41\times$ smaller compression time over the fastest baseline.
\end{abstract}

\section{Introduction}
Large language models (LLMs) have demonstrated strong capabilities when supplied with sufficiently rich context~\cite{rag}. Many complex natural language tasks require models to reason over heterogeneous sources of information, such as instructions, demonstrations, retrieved documents, conversation history, and task-specific evidence. However, existing LLMs remain constrained by finite context windows. Although substantial effort has been devoted to extending these windows, longer contexts introduce two major challenges. First, they increase inference cost and latency because the target LLM must process a larger number of input tokens~\cite{dao2022flashattention,keles2023computational}. Second, simply providing a model with more context does not necessarily improve its performance. As the input grows longer, models may struggle to effectively use relevant evidence, especially when it appears in the middle of the context~\cite{position_bias}.

Long-context compression has therefore emerged as an important approach for reducing inference cost and helping models focus on the most relevant information in long inputs. Most existing context compression methods~\cite{selective_compression,llmlingua,longllmlingua} formulate compression as a token-level pruning problem. Thus, given a long context and a compression budget, their goal is to
remove tokens estimated to be less important while retaining those required to
answer the question. A representative example is LLMLingua \cite{llmlingua}, which introduces a coarse-to-fine compression framework that uses a smaller auxiliary language model $\mathcal{M}_{S}$ to estimate the perplexity of tokens in a long context. Tokens with lower perplexity are treated as contributing less information and are removed through an iterative procedure. This approach does not consider the relevance of tokens to the question before dropping them.
LongLLMLingua \cite{longllmlingua} extends this formulation by making context compression question-aware. In particular, LongLLMLingua uses contrastive perplexity to measure how token importance changes when the context is conditioned on the question, thereby encouraging the compressor $\mathcal{M}_{S}$ to retain question-relevant information. 

\begin{figure*}[t]
\centering
\includegraphics[width=\textwidth]{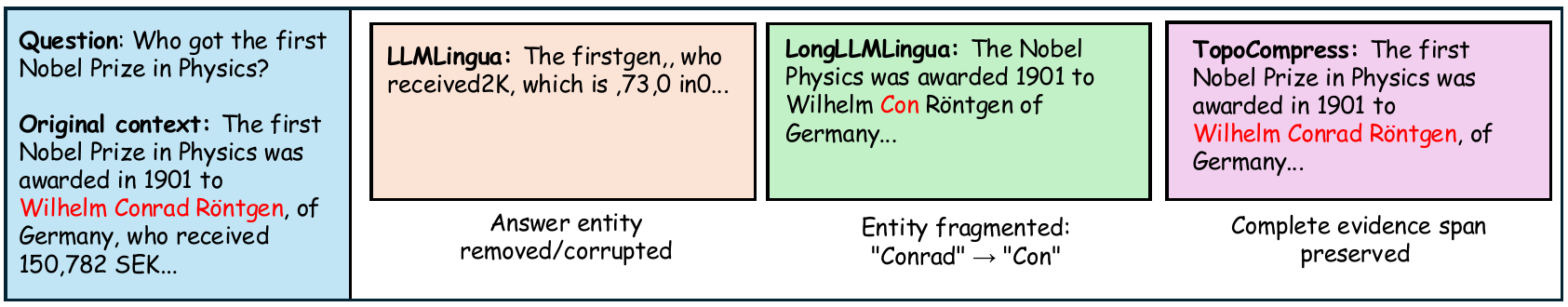}
\caption{
Example of entity fragmentation under token-level prompt compression. 
LLMLingua removes and corrupts the answer-bearing phrase, while LongLLMLingua preserves only a fragmented form. 
\texttt{TopoCompress} preserves the complete evidence.
}
\label{fig:entity_fragmentation}
\end{figure*}

While these methods demonstrate the usefulness of context compression, their
token-level perplexity-based formulation introduces several operational
limitations. First, token-level pruning can fragment the context (Figure  \ref{fig:entity_fragmentation}). Since compression decisions are made over individual tokens, the resulting compressed context may break word boundaries, phrases, named entities, or syntactic units. This can distort the information seen by the target model and may require post-hoc recovery \cite{longllmlingua}, which adds extra computation. A more principled compression method should avoid creating such fragmentation in the first place.

Second, these methods are not model-agnostic. The compressor
$\mathcal{M}_{S}$ decides which information should be removed, and the
target model $\mathcal{M}_{T}$ then uses the resulting compressed context during
inference to answer the query. $\mathcal{M}_{S}$ must learn to imitate what
$\mathcal{M}_{T}$ would consider informative in order to make reliable
compression decisions.
Existing context compression methods \cite{llmlingua,longllmlingua} attempt to reduce this mismatch through
instruction-tuning, where $\mathcal{M}_{S}$ is
adapted using responses generated by $\mathcal{M}_{T}$. However, this alignment introduces additional computation and ties the compressor to a specific target model, limiting its generality across different LLMs.

Third, perplexity-based importance scores do not account for how pieces of evidence relate to one another. Existing compression methods \cite{selective_compression, llmlingua, longllmlingua} use perplexity to estimate how predictable each token is under a small language model $\mathcal{M}_{S}$. Yet even when made question-aware, this formulation scores tokens individually rather than modeling how one piece of evidence supports or connects to another. Many multi-document and multi-hop tasks require the model to connect evidence across documents and follow causal or temporal relations before the query can be answered (Figure~\ref{fig:evidence_chain_analysis}). A token may therefore be easy to predict from its local context, and thus receive low perplexity and be dropped, while still carrying intermediate information needed to preserve the evidence chain. This makes perplexity an insufficient signal for determining which information is truly important for a downstream task.


These limitations suggest that long-context compression should move beyond treating tokens as independent units whose importance can be determined solely by a compressor model $\mathcal{M}_{S}$. Instead, compression should preserve coherent evidence, identify information that is relevant to the task, and account for relationships among different pieces of evidence. This naturally calls for a graph-based formulation, where the context is represented as a set of connected semantic units. Such a representation allows compression decisions to consider not only how relevant a piece of evidence is to the query, but also how it supports, connects, or provides intermediate information that links other evidence. Graph-based representations have been successfully used to capture relationships among textual units in applications such as summarization~\cite{bugueno2025graphlss,leite2007extractive}, conversational analysis~\cite{montero2007semi}, and information retrieval~\cite{witschel2007multi}. For the first time, we formulate long-context compression as graph-based relevance propagation.

\label{fig:musique_evidence_chain}
Based on this formulation, we introduce \texttt{TopoCompress}, a training-free and model-agnostic framework for context compression. \texttt{TopoCompress} first segments the context into coherent semantic spans and scores them based on query relevance. It further incorporates semantic acceleration to highlight query-relevant spans that occur at informative transitions in the context. The spans are then connected through a hybrid graph based on semantic similarity and sequential adjacency. Rather than selecting evidence solely from its direct relevance to the query, \texttt{TopoCompress} propagates relevance over this graph, allowing intermediate, supporting spans to gain importance through their connections to other evidence. This is useful for complex reasoning tasks, where some necessary information may not directly match the query but is essential for completing the evidence chain (Appendix~\ref{appendix:evidence_chain_analysis}).

We evaluate \texttt{TopoCompress} on five long-context question answering tasks—HotpotQA, 2WikiMQA, MuSiQue, Qasper, and MultiFieldQA-en—across several target models. The results show that \texttt{TopoCompress} achieves a strong balance between answer quality and compression efficiency, consistently outperforming all evaluated compression baselines across different compression budgets and target models. It also keeps the compression process lightweight and fast.

Our contributions are summarized as follows:
\begin{itemize}
  \item We formulate long-context compression as relevance propagation over a hybrid graph. To the best of our knowledge, this graph-relevance formulation is the first of its kind for long-context compression. 
    \item We introduce \texttt{TopoCompress}, a training-free and model-agnostic compression framework that selects compact, coherent, and non-redundant evidence by propagating query relevance over a hybrid graph, allowing evidence that may not directly match the query to be identified through its connections.
    \item We show on several long-context tasks that \texttt{TopoCompress} provides competitive compression quality while being substantially faster.
\end{itemize}

\section{Related Work}

\paragraph{Long-context Compression.}
Natural language contains redundancy~\cite{shannon1951prediction} that may aid human understanding but may be unnecessary for LLMs. Motivated by this observation, prompt compression has emerged as an effective approach for reducing the computational cost of LLM inference by removing redundant information before the input is processed by the target model. For example, Selective Context~\cite{selective_compression} uses a causal language model to estimate the self-information of lexical units and removes those considered less informative. LLMLingua~\cite{llmlingua} extends this direction by introducing a coarse-to-fine compression framework. This framework uses a smaller causal language model to estimate token importance through perplexity and iteratively prunes less informative tokens. LongLLMLingua~\cite{longllmlingua} builds on this framework with question-aware coarse-grained compression, using contrastive perplexity to prioritize information that is more relevant to the query. LLMLingua-2~\cite{llmlingua-v2} formulates prompt compression as a token classification problem. It uses data distilled from an LLM to train a bidirectional Transformer encoder that predicts whether each token should be retained or removed at inference time.

\paragraph{Graph-Based Modeling of Text.}

Graph representations have a long history in NLP as a means of capturing relationships among textual units~\cite{nastase2015survey}. TextRank~\cite{textrank} represents words and sentences as nodes in a graph and applies graph-based ranking to identify  keywords and sentences for tasks such as keyword extraction and extractive summarization. Similarly, LexRank~\cite{erkan2004lexrank} constructs a sentence-similarity graph and uses graph centrality to identify sentences for multi-document summarization. More recently, GraphLSS~\cite{bugueno2025graphlss} builds a heterogeneous graph that combines structural and semantic relations to identify salient sentences for long-document summarization.

Our work brings this graph-based perspective to long-context compression by representing context as connected semantic spans to preserve intermediate, supporting information. By doing so, it avoids reliance on a separate compressor language model or target-model-specific distillation, resulting in a training-free and model-agnostic framework for structured evidence selection.
\section{TopoCompress}
\begin{figure*}[t]
\centering
\includegraphics[width=\textwidth]{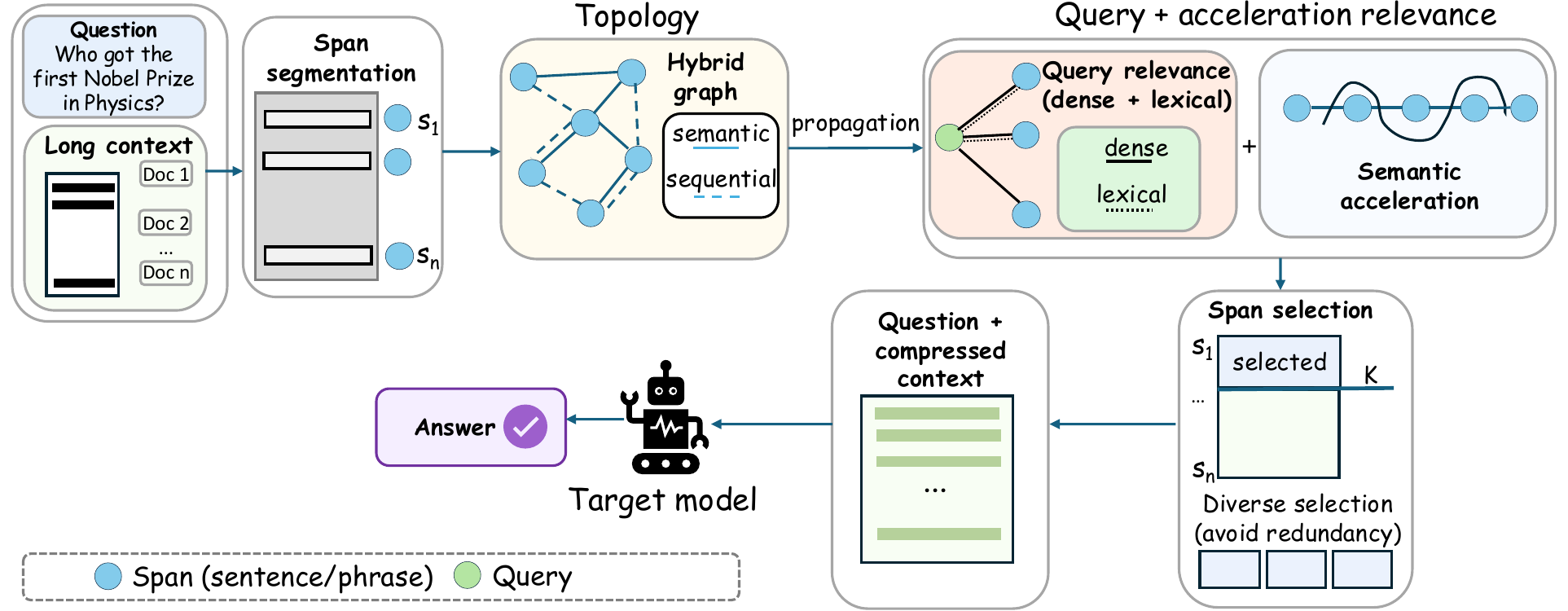}
\caption{
\texttt{TopoCompress} compresses long contexts in four main steps: (i) the context is segmented into coherent semantic spans; (ii) each span is scored using dense and lexical query relevance together with semantic acceleration; (iii) a hybrid graph is constructed using semantic similarity and sequential adjacency, and relevance is propagated over the graph; and (iv) a diverse set of high-relevance spans is selected under the compression budget and passed to the target model.}
\label{fig:topocompress_procedure}
\end{figure*}

In this section, we introduce \texttt{TopoCompress}, a training-free and model-agnostic framework for long-context compression. \texttt{TopoCompress} represents the context as connected semantic spans, scores each span using query relevance and semantic acceleration, propagates relevance through the hybrid graph, and selects a compact, non-redundant evidence set (Figure~\ref{fig:topocompress_procedure}). Algorithm~\ref{alg:topocompress} summarizes the overall procedure.

\begin{algorithm}[ht]
\caption{\texttt{TopoCompress}}
\label{alg:topocompress}
\begin{algorithmic}[1]
\REQUIRE Context documents $\mathcal{D}$, query $q$, budget $K$, frozen encoder $\Phi$
\ENSURE Compressed context $\mathcal{X}_{\mathrm{comp}}$
\STATE Segment documents into semantic spans $\mathcal{U}=\{u_1,\ldots,u_N\}$
\STATE Compute span embeddings $\mathbf{e}_i=\Phi(u_i)$ and query embedding $\mathbf{e}_q=\Phi(q)$
\STATE Compute the dense relevance score $\hat{r}_i$ using Equation~\eqref{Equaton:r} and the lexical relevance score $\hat{\ell}_i$ using Equation~\eqref{Equation:l}.
\STATE Compute query alignment $g_i=\mu\hat{r}_i+(1-\mu)\hat{\ell}_i$
\vspace{-1.2em}
\STATE Compute the semantic acceleration score $\hat{\kappa}_i$ over the original span sequence using Equation~\eqref{Equation:kappa}.
\STATE Compute the initial scores $s_i=(1+\lambda\hat{\kappa}_i)g_i$
\STATE Build span graph $\mathbf{W}=\alpha \mathbf{W}^{\mathrm{sem}}+\beta \mathbf{W}^{\mathrm{seq}}$
\STATE Normalize $\mathbf{W}$ to obtain $\mathbf{P}=\mathbf{L}^{-1}\mathbf{W}$
\STATE Propagate relevance with $\mathbf{s}^{(m+1)}=(1-\eta)\mathbf{s}+\eta \mathbf{P}^\top \mathbf{s}^{(m)}$
\STATE Greedily select non-redundant spans under budget $K$
\STATE Restore selected spans to original order
\RETURN $\mathcal{X}_{\mathrm{comp}}$
\end{algorithmic}
\end{algorithm}

\subsection{Problem Formulation}

Given a query $q$ and a context consisting of $M$ documents,
\(
\mathcal{D}=\{D_1,D_2,\ldots,D_M\},
\) the objective is to select a compact evidence set $\mathcal{X}_{\mathrm{comp}}$ that contains sufficient information to answer $q$ while satisfying a global token budget $K$:
\(
\sum_{u_i\in \mathcal{X}_{\mathrm{comp}}} \operatorname{tokens}(u_i) \le K.
\)

\subsection{Semantic Span Construction}

Token-level pruning can fragment words, entities, phrases, or relations, potentially distorting important evidence and leading the target model to incorrect conclusions. We therefore segment each document into contiguous semantic units, such as sentences or short paragraphs. Let the resulting sequence of spans be
\(
\mathcal{U}=\{u_1,u_2,\ldots,u_N\}.
\)

\subsection{Dense and Lexical Query Relevance}

Let $\Phi$ denote a frozen text encoder. We instantiate the frozen encoder $\Phi$ with \texttt{bge-m3}~\cite{m3embedding}, which is well suited for our setting because it provides both dense semantic embeddings and lexical token-weight values. For each span $u_i$, we compute a dense vector representation
\(
\mathbf{e}_i=\Phi(u_i),
\)
and the dense query representation 
\(
\mathbf{e}_q=\Phi(q).
\)

We first measure how semantically relevant each span is to the query. We refer to this as \textit{dense relevance}, since it is computed from the dense vector representations of the span and the query. Specifically, the dense relevance score of span $u_i$ is given by their cosine similarity:
\begin{equation}
\label{Equaton:r}
    r_i=\max(0,\operatorname{sim}(\mathbf{e}_i,\mathbf{e}_q)),
\end{equation}
where
\(
\operatorname{sim}(\mathbf{a},\mathbf{b})
=
\frac{\mathbf{a}^{\top}\mathbf{b}}
{\|\mathbf{a}\|_2\|\mathbf{b}\|_2}.
\)
We normalize the dense relevance scores across the candidate spans to obtain $\hat{r}_i$.

While dense relevance captures semantic similarity to the query, it may overlook exact lexical signals such as named entities, dates, or relation terms that may be important for answering the question. We therefore complement it with \textit{lexical relevance}. Let $w_t^q$ denote the lexical weight assigned by \texttt{bge-m3} to token $t$ in the query, and let $w_t^{u_i}$ denote its lexical weight in span $u_i$. We define the lexical relevance score as
\begin{equation}
    \label{Equation:l}
    \ell_i=\sum_{t\in q\cap u_i} w_t^q w_t^{u_i}.
\end{equation}
The lexical scores are also normalized across the candidate spans to obtain $\hat{\ell}_i\in[0,1]$. Finally, we combine dense and lexical relevance into a query-alignment score:
\begin{equation}
\label{eqn:g}
    g_i=\mu\hat{r}_i+(1-\mu)\hat{\ell}_i,
\end{equation}
where $\mu\in[0,1]$ controls the balance between semantic relevance and exact lexical evidence.

\subsection{Semantic Acceleration}
The query relevance information in Equation~\eqref{eqn:g} identifies spans that directly match the question, but some important spans may be useful because they occur near semantic transitions in the context. These spans may connect different pieces of evidence or mark shifts between related facts. To capture this, \texttt{TopoCompress} computes a second-order semantic acceleration score over the original span sequence.

For an interior span $u_i$ within the same document, we compute
\(
\mathbf{a}_i=2\mathbf{e}_i-\mathbf{e}_{i-1}-\mathbf{e}_{i+1},
\)
and define its acceleration magnitude as
\begin{equation}
    \label{Equation:kappa}
    \kappa_i=\|\mathbf{a}_i\|_2.
\end{equation}
The score $\kappa_i$ measures how sharply the semantic trajectory changes around span $u_i$. We normalize the acceleration scores across spans to obtain $\hat{\kappa}_i$.

Semantic acceleration should not make a query-irrelevant span important. We therefore gate acceleration with query alignment:
\begin{equation}
\label{eqn:query-gated-score}
    s_i=(1+\lambda\hat{\kappa}_i)g_i,
\end{equation}
where $\lambda\ge0$ controls the contribution of acceleration. 
This formulation boosts spans that are both query-aligned and located near strong semantic transitions, while preventing unrelated transitions from dominating the selection process.

\subsection{Span Topology and Relevance Propagation}

Some evidence needed for complex reasoning tasks may not appear highly relevant to the query in isolation. A span can instead be important because it supports, connects, or disambiguates other evidence. To capture these relationships, we build a topology over the spans and propagate the initial query-gated score vector
$\mathbf{s}=[s_1,s_2,\ldots,s_N]^\top$
(Equation~\eqref{eqn:query-gated-score}) across it.

Specifically, we construct a weighted graph $\mathcal{G}=(\mathcal{V},\mathcal{E})$, where each node $v_i$ corresponds to a span $u_i$. The graph combines \textit{semantic} and \textit{sequential} edges. For each span $u_i$, let $\mathcal{N}_k(i)$ denote the set of its $k$ nearest neighbors based on dense embedding similarity. We add semantic edges from $u_i$ to the spans in $\mathcal{N}_k(i)$, with edge weights defined as
\[
\mathbf{W}^{\mathrm{sem}}_{ij}=
\begin{cases}
\max(0,\operatorname{sim}(\mathbf{e}_i,\mathbf{e}_j)), & j\in \mathcal{N}_k(i),\\
0, & \text{otherwise}.
\end{cases}
\]

The sequential edges connect adjacent spans within the same document. Let $\operatorname{D}(u_i)$ denote the document containing span $u_i$. We define the sequential edge weight between spans $u_i$ and $u_j$ as
\[
\mathbf{W}^{\mathrm{seq}}_{ij}=
\begin{cases}
1, & |i-j|=1 \ \text{and}\ \operatorname{D}(u_i)=\operatorname{D}(u_j),\\
0, & \text{otherwise}.
\end{cases}
\]
We combine the two adjacency matrices as
\(
\mathbf{W}=\alpha \mathbf{W}^{\mathrm{sem}}+\beta \mathbf{W}^{\mathrm{seq}},
\)
where $\alpha,\beta\ge0$ control the relative contributions of semantic and sequential connectivity. We then compute the degree matrix $\mathbf{L}$ with diagonal entries
\(
\mathbf{L}_{ii}=\sum_{j=1}^N \mathbf{W}_{ij},
\) where $\qquad
\mathbf{L}_{ij}=0$ $\text{ for } i\neq j.$
We row-normalize $\mathbf{W}$ to obtain the transition matrix
\(
\mathbf{P}=\mathbf{L}^{-1}\mathbf{W}.
\)
The entry $\mathbf{P}_{ij}$ represents the proportion of relevance passed from span $u_i$ to span $u_j$ during propagation.

To allow each span to gain importance from connected evidence rather than relying solely on its direct relevance to the query, we perform personalized relevance propagation:
\[
\mathbf{s}^{(m+1)}=(1-\eta)\mathbf{s}+\eta \mathbf{P}^{\top}\mathbf{s}^{(m)},
\]
where $\eta\in(0,1)$ is the damping factor and $\mathbf{s}^{(0)}=\mathbf{s}$ is the query-gated score defined in Equation~\eqref{eqn:query-gated-score}. The restart term $(1-\eta)\mathbf{s}$ keeps the propagated scores anchored to the original query-gated signal. The update is repeated until convergence, producing the propagated score vector $\tilde{\mathbf{s}}=[\tilde{s}_1,\tilde{s}_2,\ldots,\tilde{s}_N]^\top$. This allows intermediate, supporting spans that may not directly match the query to gain importance through their connections to other relevant evidence.

\subsection{Redundancy-Aware Evidence Selection}

After propagation, spans with related meanings may receive similar scores. Selecting spans purely by $\tilde{s}_i$ can therefore introduce redundant evidence. To reduce this redundancy, \texttt{TopoCompress} selects spans greedily by penalizing candidates that are too similar to evidence already selected.

Let $\mathcal{A}$ denote the set of selected spans and let $\mathcal{R}$ denote the remaining candidate spans. At each selection step, the next span is chosen as
\[
u^*=
\arg\max_{u_i\in\mathcal{R}}
\left[
\tilde{s}_i
-
\delta
\max_{u_j\in\mathcal{A}}
\operatorname{sim}(\mathbf{e}_i,\mathbf{e}_j)
\right],
\]
where $\delta\ge0$ controls the strength of the redundancy penalty. The first term favors spans with high propagated relevance, while the second discourages selecting a span whose semantic content is already covered by the selected evidence. After selecting $u^*$, we add it to $\mathcal{A}$ and remove it from $\mathcal{R}$. The selection process continues until the global budget $K$ is reached. Finally, selected spans are restored to their original order, preserving readability and discourse coherence for the target model.

\section{Experiments}

\subsection{Models and Datasets}

\paragraph{Models.}
 We evaluate context compression using closed-source target model, GPT-5-mini~\cite{openai2025gpt5mini}, and open-weight target models, Llama-3.1-8B~\cite{llama3} and Qwen3-8B~\cite{qwen3}.  These models differ in architecture, training pipeline, and tokenization.
 
\paragraph{Datasets.}
We evaluate \texttt{TopoCompress} on long-context question answering tasks from LongBench~\cite{longbench}. LongBench is a bilingual benchmark for long-context understanding that includes tasks from single-document and multi-document settings. In this work, we focus on five QA tasks: HotpotQA, 2WikiMQA, MuSiQue, Qasper, and MultiFieldQA-en. We select these tasks because they require models to identify and reason over evidence distributed across long contexts, directly testing the ability of \texttt{TopoCompress} to preserve intermediate, supporting information under compression. HotpotQA, 2WikiMQA, and MuSiQue are multi-document QA tasks that require the model to reason over evidence distributed across multiple documents. Qasper is a scientific document QA task based on NLP papers, while MultiFieldQA-en evaluates question answering over long articles from diverse domains. These datasets have average input lengths of 9,151 words for HotpotQA, 4,887 words for 2WikiMQA, 11,214 words for MuSiQue, 3,619 words for Qasper, and 4,559 words for MultiFieldQA-en.

\subsection{Results and Discussion}
In this section, we evaluate the effectiveness and efficiency of \texttt{TopoCompress} against three strong long-context compression baselines: LLMLingua~\cite{llmlingua}, LongLLMLingua~\cite{longllmlingua}, and LLMLingua-2~\cite{llmlingua-v2}. Together, these baselines cover perplexity-based, query-aware, and token-classification approaches to context compression.

\paragraph{Performance.}
Following the evaluation setting used in LongBench, we report F1 for all five tasks. We also report the sample-weighted average F1 across the five tasks. Each compression method is evaluated under three context budgets, $K \in \{2000, 1000, 500\}$. The context budget specifies the maximum number of tokens allowed in the compressed context. The $K=500$ setting is the most aggressive compression regime, requiring each method to retain only limited evidence while preserving enough information for the target model to answer the query correctly.

\begin{figure*}[t]
\centering
\includegraphics[width=0.95\textwidth]{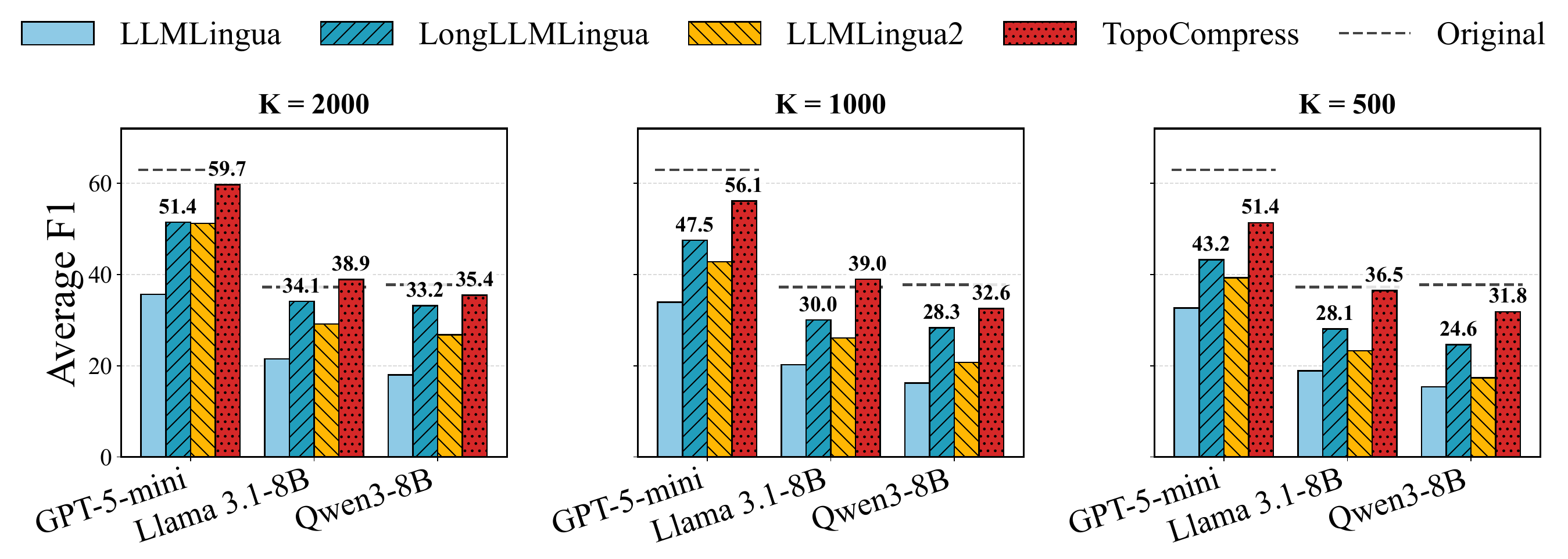}
\caption{Average F1 comparison on long context tasks across compression budgets and target models. Bars show compressed-context performance for LLMLingua, LongLLMLingua, LLMLingua-2, and \texttt{TopoCompress} at $K \in \{500,1000,2000\}$. Dashed horizontal lines indicate the corresponding full-context Original performance for each target model.}
\label{fig:longbench_avg_f1_comparison}
\end{figure*}

Figure~\ref{fig:longbench_avg_f1_comparison} presents the overall comparison of the compression methods across GPT-5-mini, Llama-3.1-8B, and Qwen3-8B. \texttt{TopoCompress} consistently achieves the best performance across different context budgets. At the most aggressive budget, $K=500$, it outperforms the strongest baseline, LongLLMLingua, by 8.15 points on GPT-5-mini, 8.44 points on Llama-3.1-8B, and 7.21 points on Qwen3-8B. Notably, the performance of \texttt{TopoCompress} at $K=500$ is comparable to that of LongLLMLingua at $K=2000$ (e.g., 51.37 vs. 51.44 on GPT-5-mini). This shows that \texttt{TopoCompress} can achieve similar performance to the strongest baseline while using a compression budget that is up to $4\times$ smaller.

Tables~\ref{tab:gpt5mini_longbench_f1}, \ref{tab:llama31_8b_longbench_f1}, and \ref{tab:qwen3_longbench_f1} in Appendix~\ref{appendix:additional-topocompress} provide detailed task-level results for GPT-5-mini, Llama-3.1-8B, and Qwen3-8B, respectively. Under aggressive compression, \texttt{TopoCompress} achieves strong gains on multi-document QA tasks such as HotpotQA, 2WikiMQA, and MuSiQue, where answering a query often requires combining evidence distributed across multiple documents. These results highlight the importance of preserving connected evidence for multi-hop reasoning and show that the query-gated, graph-guided selection in \texttt{TopoCompress} effectively retains such evidence even under tight compression budgets.

\paragraph{Compression Time.}
In Table~\ref{tab:compression_speed}, we report the compression time for all compression methods at $K=1000$. \texttt{TopoCompress} is the fastest method, completing compression across all five LongBench tasks in less than 7 minutes. Compared with LLMLingua2, the fastest baseline, \texttt{TopoCompress} achieves a $1.41\times$ speedup. The difference is substantially larger compared with LLMLingua and LongLLMLingua, which require approximately 44 and 52 minutes, respectively. These results demonstrate that \texttt{TopoCompress} not only preserves higher-quality evidence for the target model but also significantly reduces compression time, making it easier to integrate into long-context pipelines such as RAG systems, where compression may be required to improve inference efficiency.

\begin{table*}[!t]
\centering
\small
\setlength{\tabcolsep}{7pt}
\begin{tabular}{lccccccc}
\toprule
\textbf{Method} & \textbf{HotpotQA} & \textbf{2WikiMQA} & \textbf{MuSiQue} & \textbf{Qasper} & \textbf{MultiFieldQA-en} & \textbf{Overall} & \textbf{Speedup} \\
\midrule
LLMLingua     & 11:52 & 7:16 & 14:02 & 5:33  & 5:16 & 43:59 & 0.22$\times$ \\
LongLLMLingua & 11:02 & 7:30 & 12:41 & 11:39 & 8:58 & 51:50 & 0.19$\times$ \\
LLMLingua2    & 2:42  & 1:29 & 3:13  & 1:10  & 1:09 & 9:43  & 1.00$\times$ \\
TopoCompress  & \textbf{1:59} & \textbf{1:04} & \textbf{2:03} & \textbf{0:53} & \textbf{0:56} & \textbf{6:54} & \textbf{1.41$\times$} \\
\bottomrule
\end{tabular}
\caption{Compression time at $K=1000$ for long-context compression methods. Times are reported in minutes:seconds. Speedup is computed relative to LLMLingua-2 overall runtime, the fastest baseline.}
\label{tab:compression_speed}
\end{table*}

\subsection{Controller for Adaptive  Reasoning \label{sec:controller}
}
\citet{itercomp} showed that LLMs can serve as missing-information identifiers: given a question and partial evidence, an LLM can determine whether the evidence is sufficient and, if not, formulate a focused query for the information still needed. This can be useful in complex settings where the original question may not explicitly reveal every intermediate clue. Motivated by this idea, we evaluate \texttt{TopoCompress} with an LLM-based controller in the loop.

After half of the evidence budget (i.e., $0.5K$) has been accumulated, the controller determines whether the selected evidence is sufficient to answer the original question. If it returns \textit{answerable}, selection stops early and the accumulated evidence is used as the compressed context. If it returns \textit{unanswerable}, the controller generates a focused follow-up question to retrieve additional evidence that may not be directly captured by the original question. \texttt{TopoCompress} then re-scores the remaining spans using this question while keeping document-side quantities, including span embeddings and semantic acceleration fixed. To limit API overhead, the controller is invoked at most twice per sample.



\subsubsection{Controller Analysis}
\label{sec:controller_analysis}

The controller variant provides a useful comparison for understanding the strength of the TopoCompress selector. We analyze its effect from two perspectives: the amount of context retained after compression and the downstream QA performance. For this experiment, we use GPT-5-mini as the controller to generate intermediate reasoning queries that guide additional evidence selection during compression.

\paragraph{Impact on Context Length.}
We report the average compressed context length for TopoCompress with and without the controller at $K=1000$ in Table~\ref{tab:controller_context_length}. Avg. Context is the average token length of the compressed evidence context given to the target model, excluding the task instruction and question. It directly measures how much source context remains after compression. In the standard TopoCompress setting, the method typically fills close to the budgeted context length. With the controller enabled,  the average context tokens across all tasks reduces from 1000 tokens to roughly 681--857 tokens depending on the task. 

\begin{table}[t]
\centering
\small
\setlength{\tabcolsep}{6pt}
\begin{tabular}{lcc}
\toprule
\textbf{Task} & \textbf{Avg. Ctx.} & \textbf{Avg. Ctx. + Ctrl.} \\
\midrule
HotpotQA        & 1000.0 & 729.2 \\
2WikiMQA        & 999.0  & 803.3 \\
MuSiQue         & 1001.0 & 857.1 \\
Qasper          & 998.1  & 734.5 \\
MultiFieldQA-en & 999.2  & 680.9 \\
\bottomrule
\end{tabular}
\caption{Average compressed context (Ctx) length at $K=1000$ with and without the controller (Ctrl).}
\label{tab:controller_context_length}
\end{table}

\paragraph{Impact on Performance.}
As shown in Table~\ref{tab:controller_experiment}, the controller variant slightly improves performance over the controller-free \texttt{TopoCompress} setting. This suggests that the original question may not always surface all the evidence needed to answer complex problems, and that constructing intermediate reasoning queries can help recover additional supporting evidence. However, the gains are modest, indicating that \texttt{TopoCompress} already captures much of this benefit through its query-guided and graph-based selection mechanism. In other words, \texttt{TopoCompress} can trace and recover much of the relevant intermediate evidence without requiring an LLM in the compression loop.

\subsection{Ablation Study}
\label{sec:ablation}

We conduct an ablation study to assess the contribution of the  scoring components in TopoCompress. We evaluate the ablations at two compression budgets, $K=2000$ and $K=500$.

\paragraph{TopoCompress without graph propagation (\textit{w/o Graph}).}

Graph propagation diffuses query relevance through the hybrid span topology, allowing \texttt{TopoCompress} to identify evidence that may not be selected by direct query matching but is connected to relevant evidence through semantic similarity or sequential adjacency. To assess the full contribution of graph propagation, we focus this ablation on multi-hop and multi-document tasks, where answering questions depends on preserving connected evidence chains.

As shown in Table~\ref{tab:graph_ablation_multihop} (Appendix~\ref{appendix:ablation}), removing graph propagation consistently reduces performance on HotpotQA, 2WikiMQA, and MuSiQue across different target models and compression budgets. The relative drops reach 14.1\% on MuSiQue, 7.4\% on HotpotQA, and 5.9\% on 2WikiMQA. These results indicate that graph-based relevance propagation helps preserve connected evidence under compression, particularly for tasks that require reasoning over multiple supporting spans.

\paragraph{Additional ablations.}

We further evaluate two component removals in \texttt{TopoCompress}: without query relevance (\textit{w/o Query Rel.}), which removes the dense--lexical query relevance signal, and without semantic acceleration (\textit{w/o Accel.}), which removes the semantic acceleration term. We conduct these ablations using GPT-5-mini as the target model (Table~\ref{tab:ablation_gpt} in Appendix~\ref{appendix:ablation}).

The results show that removing query relevance causes the largest performance drop. This shows that the dense--lexical query gate is essential for identifying spans that are directly relevant to answering the question. However, multi-hop tasks may also require intermediate evidence that does not directly match the query (Appendix~\ref{appendix:evidence_chain_analysis}). Our topology-aware refinement propagates query relevance through the graph to account for such connections. The semantic acceleration component has a comparatively smaller impact. Removing acceleration changes the average F1  modestly.



\section{Conclusion}

We introduced \texttt{TopoCompress}, a training-free and model-agnostic framework that formulates long-context compression as graph-structured evidence selection over coherent semantic spans. Our method combines dense--lexical query relevance, semantic acceleration, and graph-based propagation over semantic and sequential span connections to select compact, non-redundant evidence for the target model. Our findings show that \texttt{TopoCompress} consistently outperforms strong context compression baselines across both closed-source and open-weight target models. In addition to these performance gains, \texttt{TopoCompress} is faster than existing baselines, making it practical for long-context pipelines. These results establish graph-based evidence selection as a more effective and practical paradigm for long-context compression, moving beyond token-level pruning and target-model-dependent designs.

\bibliography{references}

\appendix

\section{Graph-Based Evidence Recovery}
\label{appendix:evidence_chain_analysis}

\begin{figure*}[htb]
\centering
\includegraphics[width=0.9\textwidth]{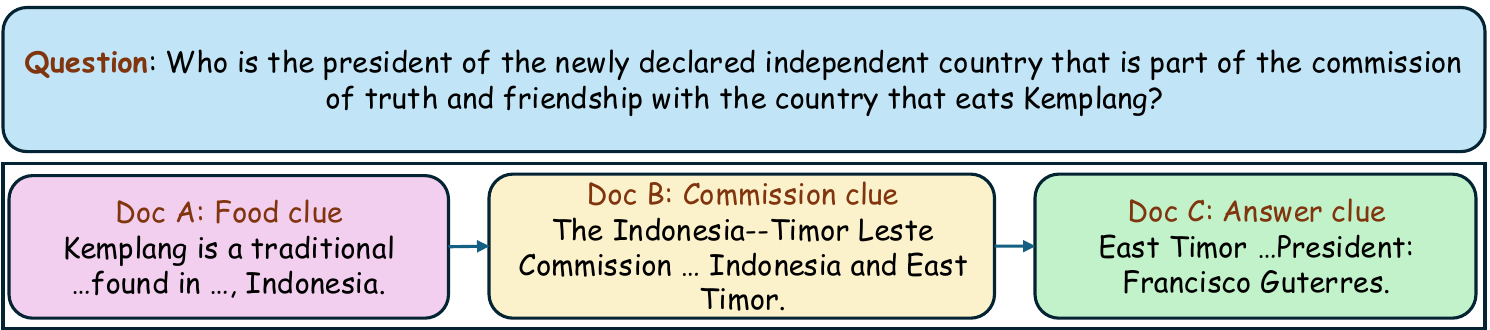}
\caption{
Evidence-chain illustration of multi-hop reasoning. Answering the question requires following intermediate links from Kemplang to Indonesia, from Indonesia to East Timor, and from East Timor to its president.
}
\label{fig:evidence_chain_analysis}
\end{figure*}

Multi-hop questions often require connecting evidence across several documents rather than selecting a single directly matched span. However, an answer-bearing document may only become identifiable through intermediate evidence. Figure~\ref{fig:evidence_chain_analysis} illustrates this setting, where the question requires following a chain from one piece of evidence to another. In such situations, using query relevance alone may be insufficient to retrieve the answer-bearing evidence.

We illustrate how graph propagation can recover evidence that is missed by direct query relevance. We conduct this analysis using an example from the MuSiQue dataset~\cite{trivedi2022musique}. MuSiQue provides gold annotations for the supporting paragraphs required to answer each question, which allows us to inspect whether the ranked spans correspond to the true evidence chain. We denote each span as \texttt{p\_i:span\_j}, where \texttt{p\_i} indicates the $i$-th paragraph in the input context and \texttt{span\_j} indicates the $j$-th span produced after segmenting that paragraph. 

As shown in Table~\ref{tab:musique_graph_case_study}, query relevance alone can assign high scores to spans that are not part of the gold evidence chain, causing them to appear near the top of the ranking. For example, a non-supporting span is ranked 3rd by query relevance, while graph propagation move it to rank 11. More importantly, the answer-bearing span \texttt{p\_19:span\_3} is ranked only 14th by query relevance and would therefore be excluded under a tight budget. After graph propagation, this span is promoted to rank 7, allowing the compressed context to retain the evidence needed to answer the question. This illustrates the role of graph propagation beyond direct query matching: it helps recover connected supporting evidence that may be weakly matched to the question itself but is topologically linked to other relevant evidence.


\begin{table*}[t]
\centering
\scriptsize
\setlength{\tabcolsep}{4pt}
\renewcommand{\arraystretch}{1.15}
\begin{tabular}{p{2.2cm}p{9.4cm}cc}
\toprule
\multicolumn{4}{p{15.2cm}}{
\textbf{Question:} Who is the president of the newly declared independent country that is part of the commission of truth and friendship with the country that eats Kemplang?
} \\
\midrule
\textbf{Passage / Span} & \textbf{Span Text} & \textbf{Query Rank} & \textbf{Graph Rank} \\
\midrule
\rowcolor{gray!15}
p\_0:span\_0 &
Kemplang is a traditional savory fish cracker commonly found in southern parts of Sumatra, Indonesia. &
1 & 9 \\

p\_9:span\_1 &
He was that country's last governor before independence, from 1968 to 1975, and first president after it gained independence from the Netherlands. &
3 & 11 \\

\rowcolor{gray!15}
p\_3:span\_2 &
After holding private hearings and document reviews, the commission handed in the final report to the presidents of both nations and was endorsed by Indonesian President Susilo Bambang Yudhoyono. &
4 & 2 \\

\rowcolor{gray!15}
p\_3:span\_1 &
The commission was officially created to investigate acts of violence that occurred around the independence referendum held in East Timor in 1999. &
9 & 5 \\

\rowcolor{gray!15}
p\_3:span\_0 &
The Indonesia--Timor Leste Commission on Truth and Friendship was established jointly by the governments of Indonesia and East Timor in August 2005. &
11 & 10 \\

\rowcolor{gray!15}
p\_19:span\_3 &
Government: unitary semi-presidential constitutional republic. President: Francisco Guterres. Prime Minister: Mari Alkatiri. &
14 & 7 \\
\bottomrule
\end{tabular}
\caption{
Qualitative analysis of multi-hop evidence recovery on the MuSiQue dataset using graph propagation. MuSiQue provides gold annotations for the documents or paragraphs that contain the clues needed to answer each question. Rows shaded in gray indicate spans derived from these gold supporting paragraphs.
}
\label{tab:musique_graph_case_study}
\end{table*}

\section{Additional Results for \texttt{TopoCompress}
\label{appendix:additional-topocompress}}

Additional results for \texttt{TopoCompress} and the evaluated baselines are shown in Table~\ref{tab:gpt5mini_longbench_f1} for GPT-5-mini, Table~\ref{tab:llama31_8b_longbench_f1} for Llama-3.1-8B, and Table~\ref{tab:qwen3_longbench_f1} for Qwen3-8B.

\begin{table*}[t]
\centering
\scriptsize
\setlength{\tabcolsep}{5pt}
\begin{tabular}{lccccccc}
\toprule
\textbf{Method} & \textbf{$K$} & \textbf{HotpotQA} & \textbf{2WikiMQA} & \textbf{MuSiQue} & \textbf{Qasper} & \textbf{MultiFieldQA-en} & \textbf{Avg. F1} \\
\midrule
Original & -- & 71.64 & 80.74 & 63.37 & 45.61 & 49.97 & 62.92 \\
\midrule
LLMLingua     & 2000 & 48.45 & 46.14 & 32.42 & 24.15 & 24.16 & 35.64 \\
LongLLMLingua & 2000 & 67.30 & 71.88 & 46.75 & 32.36 & 34.76 & \underline{51.44} \\
LLMLingua2    & 2000 & 58.89 & 70.04 & 42.93 & 42.91 & 38.13 & 51.24 \\
TopoCompress  & 2000 & 68.71 & 78.23 & 54.72 & 44.08 & 50.37 & \textbf{59.69} \\
\midrule
LLMLingua     & 1000 & 48.24 & 45.13 & 33.08 & 21.00 & 18.28 & 33.93 \\
LongLLMLingua & 1000 & 67.03 & 65.43 & 44.06 & 25.45 & 31.49 & \underline{47.49} \\
LLMLingua2    & 1000 & 50.89 & 53.67 & 38.03 & 36.87 & 31.57 & 42.77 \\
TopoCompress  & 1000 & 66.73 & 70.03 & 49.06 &  42.67 & 50.82 & \textbf{56.13} \\
\midrule
LLMLingua     & 500 & 49.01 & 44.84 & 28.14 & 18.15 & 20.00 & 32.66 \\
LongLLMLingua & 500 & 62.89 & 58.24 & 40.42 & 19.88 & 31.85 & \underline{43.22} \\
LLMLingua2    & 500 & 50.63 & 48.11 & 32.32 & 35.87 & 26.08 & 39.26 \\
TopoCompress  & 500 & 63.14 & 63.25 & 44.96 & 38.04 & 46.17 & \textbf{51.37} \\
\bottomrule
\end{tabular}
\caption{F1 results of different long-context compression methods with GPT-5-mini as the target model. $K$ denotes the maximum number of tokens in the compressed context. Avg. F1 is computed as a sample-weighted average across tasks. Bold indicates the best compressed method by Avg. F1 at each budget, while \underline{underline} indicates the second-best.
}
\label{tab:gpt5mini_longbench_f1}
\end{table*}

\begin{table*}[t]
\centering
\scriptsize
\setlength{\tabcolsep}{5pt}
\begin{tabular}{lccccccc}
\toprule
\textbf{Method} & \textbf{$K$} & \textbf{HotpotQA} & \textbf{2WikiMQA} & \textbf{MuSiQue} & \textbf{Qasper} & \textbf{MultiFieldQA-en} & \textbf{Avg. F1} \\
\midrule
Original & -- & 44.17 & 34.37 & 31.90 & 29.15 & 49.96 & 37.28 \\
\midrule
LLMLingua     & 2000 & 28.85 & 24.55 & 14.35 & 17.38 & 22.70 & 21.51 \\
LongLLMLingua & 2000 & 48.09 & 38.74 & 28.97 & 21.65 & 32.67 & \underline{34.10} \\
LLMLingua2    & 2000 & 35.41 & 30.14 & 21.84 & 23.08 & 37.27 & 29.14 \\
TopoCompress  & 2000 & 52.01 & 37.70 & 32.50 & 27.07 & 47.24 & \textbf{38.89} \\
\midrule
LLMLingua     & 1000 & 29.95 & 25.84 & 11.10 & 16.85 & 16.40 & 20.22 \\
LongLLMLingua & 1000 & 46.46 & 30.04 & 23.26 & 20.20 & 30.18 & \underline{30.02} \\
LLMLingua2    & 1000 & 35.99 & 24.08 & 19.55 & 20.87 & 31.47 & 26.12 \\
TopoCompress  & 1000 & 52.68 & 37.07 & 31.60 & 28.14  & 47.37 & \textbf{38.95} \\
\midrule
LLMLingua     & 500 & 29.63 & 22.90 & 12.17 & 15.29 & 12.93 & 18.88 \\
LongLLMLingua & 500 & 44.60 & 28.50 & 21.38 & 17.76 & 28.09 & \underline{28.06} \\
LLMLingua2    & 500 & 28.12 & 26.38 & 15.44 & 20.19 & 27.34 & 23.29 \\
TopoCompress  & 500 & 51.69 & 37.74 & 25.08 & 25.23 & 44.83 & \textbf{36.50} \\
\bottomrule
\end{tabular}
\caption{F1 results of different long-context compression methods with Llama 3.1-8B as the target model.}
\label{tab:llama31_8b_longbench_f1}
\end{table*}

\begin{table*}[t]
\centering
\scriptsize
\setlength{\tabcolsep}{5pt}
\begin{tabular}{lccccccc}
\toprule
\textbf{Method} & \textbf{$K$} & \textbf{HotpotQA} & \textbf{2WikiMQA} & \textbf{MuSiQue} & \textbf{Qasper} & \textbf{MultiFieldQA-en} & \textbf{Avg. F1} \\
\midrule
Original & -- & 49.16 & 33.34 & 19.17 & 41.80 & 48.53 & 37.78 \\
\midrule
LLMLingua     & 2000 & 20.67 & 21.35 & 5.65  & 18.58 & 25.71 & 18.01 \\
LongLLMLingua & 2000 & 45.31 & 35.95 & 20.72 & 26.98 & 38.12 & \underline{33.17} \\
LLMLingua2    & 2000 & 32.00 & 23.81 & 8.10  & 37.29 & 34.70 & 26.78 \\
TopoCompress  & 2000 & 45.29 & 29.76 & 16.59 & 41.07 & 47.47 & \textbf{35.44} \\
\midrule
LLMLingua     & 1000 & 18.18 & 24.14 & 5.06  & 13.28 & 21.86 & 16.22 \\
LongLLMLingua & 1000 & 38.50 & 32.60 & 16.48 & 21.07 & 34.62 & \underline{28.34} \\
LLMLingua2    & 1000 & 21.03 & 18.61 & 7.75  & 28.91 & 29.64 & 20.74 \\
TopoCompress  & 1000 & 39.36 & 27.93 & 13.59 & 39.00 & 46.40 & \textbf{32.56} \\
\midrule
LLMLingua     & 500 & 18.12 & 22.82 & 4.45  & 12.31 & 20.53 & 15.39 \\
LongLLMLingua & 500 & 33.18 & 26.58 & 15.15 & 16.99 & 33.49 & \underline{24.63} \\
LLMLingua2    & 500 & 17.21 & 15.88 & 5.85  & 22.99 & 27.42 & 17.36 \\
TopoCompress  & 500 & 37.98 & 28.32 & 13.59 & 37.07 & 45.72 & \textbf{31.84} \\
\bottomrule
\end{tabular}
\caption{F1 results of different long-context compression methods with Qwen3-8B as the target model.
}
\label{tab:qwen3_longbench_f1}
\end{table*}

\section{Controller-Guided \texttt{TopoCompress}
\label{appendix:controller}
}
In this section, we provide results for the controller-guided variant of \texttt{TopoCompress} (Section~\ref{sec:controller}). The controller generates intermediate reasoning queries during compression to guide additional evidence selection when the current compressed context appears insufficient. The results are shown in Table~\ref{tab:controller_experiment}.

\begin{table*}[!t]
\centering
\scriptsize
\setlength{\tabcolsep}{4pt}
\begin{tabular}{llccccccc}
\toprule
\textbf{Target Model} & \textbf{Method} & \textbf{$K$} & \textbf{HotpotQA} & \textbf{2WikiMQA} & \textbf{MuSiQue} & \textbf{Qasper} & \textbf{MultiFieldQA-en} & \textbf{Avg. F1} \\
\midrule
\multirow{6}{*}{GPT-5-mini}
& TopoCompress & 2000 & 68.71 & 78.23 & 54.72 & 44.08 & 50.37 & 59.69 \\
& TopoCompress + Controller & 2000 & 69.60 & 76.80 & 57.68 & 43.75 & 48.77 & \textbf{59.88} \\
\cmidrule{2-9}
& TopoCompress & 1000 & 66.73 & 70.03 & 49.06 & 42.67 & 50.82 & 56.13 \\
& TopoCompress + Controller & 1000 & 67.29   & 75.04 & 56.96 & 43.40 & 48.60 & \textbf{58.77} \\
\cmidrule{2-9}
& TopoCompress & 500 & 63.14 & 63.25 & 44.96 & 38.04 & 46.17 & 51.37 \\
& TopoCompress + Controller & 500 & 65.42 & 64.67 & 49.78 & 41.91 & 49.86 & \textbf{54.56} \\
\midrule

\multirow{6}{*}{Llama 3.1-8B}
& TopoCompress & 2000 & 52.01 & 37.70 & 32.50 & 27.07 & 47.24 & 38.89 \\
& TopoCompress + Controller & 2000 & 55.45 & 42.39 & 34.46 & 28.87 & 46.72 & \textbf{41.31} \\
\cmidrule{2-9}
& TopoCompress & 1000 & 52.68 & 37.07 & 31.60 & 28.14 & 47.37 & 38.95 \\
& TopoCompress + Controller & 1000 & 52.03 & 46.64 & 35.10 & 27.64 & 43.41 & \textbf{40.83} \\
\cmidrule{2-9}
& TopoCompress & 500 & 51.69 & 37.74 & 25.08 & 25.23 & 44.83 & 36.50 \\
& TopoCompress + Controller & 500 & 51.41 & 38.52 & 29.94 & 27.21 & 44.07 & \textbf{37.92} \\
\midrule

\multirow{6}{*}{Qwen3-8B}
& TopoCompress & 2000 & 45.29 & 29.76 & 16.59 & 41.07 & 47.47 & \textbf{35.44} \\
& TopoCompress + Controller & 2000 & 41.21 & 29.59 & 18.44 & 39.78 & 46.06 & 34.43 \\
\cmidrule{2-9}
& TopoCompress & 1000 & 39.36 & 27.93 & 13.59 & 39.00 & 46.40 & 32.56 \\
& TopoCompress + Controller & 1000 & 41.13  & 31.45  & 19.28 & 38.57 & 46.11 & \textbf{34.74} \\
\cmidrule{2-9}
& TopoCompress & 500 & 37.98 & 28.32 & 13.59 & 37.07 & 45.72 & 31.84 \\
& TopoCompress + Controller & 500 & 34.89 & 32.66 & 16.78 & 37.93 & 45.37 & \textbf{32.90} \\
\bottomrule
\end{tabular}
\caption{Experiment comparing TopoCompress with and without the controller on several tasks across GPT-5-mini, Llama 3.1-8B, and Qwen3-8B. $K$ denotes the compressed context budget. Bold indicates the better Avg. F1 within each target model and budget.}
\label{tab:controller_experiment}
\end{table*}

\section{Ablation Study Results
\label{appendix:ablation}
}
The ablation results are shown in Tables~\ref{tab:graph_ablation_multihop} and~\ref{tab:ablation_gpt}.

\makeatletter
\setlength{\@dblfptop}{0pt}
\setlength{\@dblfpsep}{24pt}
\setlength{\@dblfpbot}{0pt plus 1fil}
\makeatother
\begin{table*}[!t]
\centering
\scriptsize
\setlength{\tabcolsep}{8pt}
\begin{tabular}{llcccc}
\toprule
\textbf{Target Model} & \textbf{Method} & \textbf{$K$} & \textbf{HotpotQA} & \textbf{2WikiMQA} & \textbf{MuSiQue} \\
\midrule
Llama3.1-8B & TopoCompress & 2000 & 52.01 & 37.70 & 32.50 \\
             & w/o Graph    & 2000 & 50.34 {\scriptsize(-3.2\%)} & 35.46 {\scriptsize(-5.9\%)} & 30.97 {\scriptsize(-4.7\%)} \\
\cmidrule(lr){2-6}
             & TopoCompress & 500  & 51.69 & 37.74 & 25.08 \\
             & w/o Graph    & 500  & 47.85 {\scriptsize(-7.4\%)} & 35.94 {\scriptsize(-4.8\%)} & 24.82 {\scriptsize(-1.0\%)} \\
\midrule
Qwen3-8B     & TopoCompress & 2000 & 45.29 & 29.76 & 16.59 \\
             & w/o Graph    & 2000 & 44.91 {\scriptsize(-0.8\%)} & 29.35 {\scriptsize(-1.4\%)} & 15.16 {\scriptsize(-8.6\%)} \\
\cmidrule(lr){2-6}
             & TopoCompress & 500  & 37.98 & 28.32 & 13.59 \\
             & w/o Graph    & 500  & 36.56 {\scriptsize(-3.7\%)} & 26.64 {\scriptsize(-5.9\%)} & 11.68 {\scriptsize(-14.1\%)} \\
\midrule
GPT-5 mini   & TopoCompress & 2000 & 68.71 & 78.23 & 54.72 \\
             & w/o Graph    & 2000 & 68.18 {\scriptsize(-0.8\%)} & 77.74 {\scriptsize(-0.6\%)} & 53.38 {\scriptsize(-2.4\%)} \\
\cmidrule(lr){2-6}
             & TopoCompress & 500  & 63.14 & 63.25 & 44.96 \\
             & w/o Graph    & 500  & 61.89 {\scriptsize(-2.0\%)} & 60.72 {\scriptsize(-4.0\%)} & 43.98 {\scriptsize(-2.2\%)} \\
\bottomrule
\end{tabular}

\caption{
Graph propagation ablation on multi-hop long context tasks. Values in parentheses indicate the relative F1 drop after removing graph propagation.
}
\label{tab:graph_ablation_multihop}
\end{table*}

\begin{table*}[!t]
\centering
\scriptsize
\setlength{\tabcolsep}{5pt}
\begin{tabular}{lccccccc}
\toprule
\textbf{Method} & \textbf{$K$} & \textbf{HotpotQA} & \textbf{2WikiMQA} & \textbf{MuSiQue} & \textbf{Qasper} & \textbf{MultiFieldQA-en} & \textbf{Avg. F1} \\
\midrule
TopoCompress   & 2000 & 68.71 & 78.23 & 54.72 & 44.08 & 50.37 & 59.69 \\
w/o Query Rel. & 2000 & 61.36 & 60.59 & 43.09 & 40.99 & 37.21 & 49.25 \\
w/o Accel.     & 2000 & 69.81 & 77.08 & 53.98 & 43.68 & 49.95 & 59.37 \\
\midrule
TopoCompress   & 500 & 63.14 & 63.25 & 44.96 & 38.04 & 46.17 & 51.37 \\
w/o Query Rel. & 500 & 57.65 & 44.17 & 35.83 & 27.83 & 23.73 & 38.58 \\
w/o Accel.     & 500 & 61.75 & 63.45 & 42.79 & 37.99 & 48.31 & 50.99 \\
\bottomrule
\end{tabular}
\caption{Ablation results for TopoCompress on long context tasks using GPT-5-mini as the target model.}
\label{tab:ablation_gpt}
\end{table*}

\end{document}